\documentclass[10pt,twocolumn]{article}
\usepackage{simpleConference}
\usepackage{authblk}
\usepackage{graphicx}
\usepackage{svg}
\usepackage{amssymb}
\usepackage[T1]{fontenc}
\usepackage[]{hyperref}
\usepackage{xcolor}
\usepackage[
    labelsep=period,
    labelfont=bf,
    textfont=bf
]{caption}
\usepackage{multirow}
\usepackage{booktabs}
\usepackage{float}
\usepackage{array}
\usepackage[noabbrev,nameinlink,capitalise]{cleveref}
\usepackage{csquotes}
\usepackage[
    backend=biber,
    citestyle=authoryear,
    natbib=true,
    style=authoryear,
    maxnames=2,
    maxbibnames=5
]{biblatex}
\DeclareNameAlias{sortname}{given-family}
\usepackage{scalerel,xparse}
\usepackage[framemethod=TikZ]{mdframed}
\usepackage{comment}
\usepackage[normalem]{ulem}

\usepackage{balance}

\newcommand{\pmind}{\textsc{PersianMind }}
\newcommand{\pmindns}{\textsc{PersianMind}}

\definecolor{bcprompt}{rgb}{0.85,0.89,0.95}
\definecolor{brdrprompt}{rgb}{0.56,0.66,0.86}
\newmdenv[
    roundcorner=10pt, 
    backgroundcolor=bcprompt, 
    linecolor=brdrprompt, 
    linewidth=5pt
]{prompt}

\definecolor{bcresponse}{rgb}{0.98,0.89,0.83}
\definecolor{brdrresponse}{rgb}{0.95,0.69,0.51}
\newmdenv[
    roundcorner=10pt, 
    backgroundcolor=bcresponse, 
    linecolor=brdrresponse, 
    linewidth=5pt
]{response}

\title{\pmindns: A Cross-Lingual Persian-English Large Language Model}

\author[ \hspace{-1ex}]{Pedram Rostami}
\author[ \hspace{-1ex}]{Ali Salemi}
\author[ \hspace{-1ex}]{Mohammad Javad Dousti}
\affil[ ]{
\{%
    \href{mailto:pedram.rostami@ut.ac.ir}{pedram.rostami},
    \href{mailto:alisalemi@ut.ac.ir}{alisalemi},
    \href{mailto:mjdousti@ut.ac.ir}{mjdousti}\}@ut.ac.ir
}
\affil[ ]{University of Tehran}

\begin{document}

\maketitle
\thispagestyle{empty}

\begin{abstract}
Large language models demonstrate remarkable proficiency in various linguistic tasks and have extensive knowledge across various domains.
Although they perform best in English, their ability in other languages is notable too.
In contrast, open-source models, such as LLaMa, are primarily trained on English datasets, resulting in poor performance in non-English languages.
In this paper, we introduce \pmindns, an open-source bilingual large language model which demonstrates comparable performance to closed-source GPT-3.5-turbo in the Persian language.
By expanding LLaMa2's vocabulary with 10,000 Persian tokens and training it on a dataset comprising nearly 2 billion Persian tokens, we show that our approach preserves the model's English knowledge and employs transfer learning to excel at transferring task knowledge from one language to another.

\textit{Keywords}: Large Language Model, LLaMa, Persian language
\end{abstract}

\section{Introduction}

\textit{Large language models} (LLMs) have been the most significant development in the field of natural language processing in recent years, reintroducing the concept of employing a model as an \textit{artificial general intelligence} (AGI)~\citep{bubeck2023sparks}.
These transformer-based decoder-only~\citep{vaswani2017attention} models are distinguished by their considerable scale and training on extensive textual datasets.
LLMs are versatile tools for various language tasks in natural language processing.
Furthermore, there are instances that LLM-based chatbots can replace traditional information retrieval systems~\citep{zhao2023survey}.
When combined with multi-modal capabilities, they can also be used in computer vision, as demonstrated by the emergence of visual chatbots~\citep{wu2023visual}.

Prominent commercial LLMs such as ChatGPT~\citep{openai2023chatgpt}, GPT-4~\citep{openai2023gpt4}, PaLM2~\citep{anil2023palm}, and Claude~\citep{anthropic2023claude} demonstrate excellent performance across a diverse range of tasks, including text generation, summarization, and code generation.
Additionally, these models exhibit promising results when applied to non-English languages.
However, it is important to note that these LLMs are proprietary and come with certain limitations for fine-tuning and access to their original models is restricted.
On the other hand, although open-source LLMs like LLaMa2 have demonstrated impressive results in the English language, their performance significantly degrades when applying to other languages~\citep{touvron2023llama2}.
This disparity can be attributed to the fact that their training dataset consist of mostly English texts, which limits their ability to understand and generate contents in other languages.

To address the poor performance of open-source LLMs in the Persian language, we introduce \pmindns\footnote{The model can be downloaded from \url{https://huggingface.co/universitytehran/PersianMind-v1.0}.}, an open-source Persian-English LLM that achieves comparable results to GPT-3.5-turbo~\citep{openai2023chatgpt} in reading comprehension benchmark.
In this paper, we employed a Persian Byte-Pair Encoding tokenizer comprising 10,000 tokens, which has been trained on a cleaned Persian Wikipedia corpus.
These tokens are added to LLaMa2's vocabulary, and the model's embeddings are subsequently expanded.
We utilize the LoRA technique to train our model on a 2-billion-token Persian corpus and subsequently fine-tune the model using various instruction tuning datasets to enhance its performance on natural language processing tasks.
Due to limited instruction tuning datasets for the Persian language, we refined our model by fine-tuning it on high-quality Persian machine-translated datasets.

Our key contributions in this paper are as follows:

\begin{itemize}
    \item[$\bullet$] Introduced \pmindns, an open-source Persian-English large language model trained using a cost-aware approach which utilizes the LoRA technique and data parallelism.

    \item[$\bullet$] Achieved state-of-the-art results on Persian subset of the Belebele benchmark and the ParsiNLU multiple-choice QA task.
    \item[$\bullet$] Attained performance comparable to GPT-3.5-turbo in a Persian reading comprehension task.
    \item[$\bullet$] Alleviated catastrophic forgetting resulting from extensive training on Persian datasets by training on Persian-English parallel datasets and employing the LoRA technique.
    \item[$\bullet$] Demonstrated that \pmind can generate high-quality sentence embeddings, surpassing the performance of previous masked language models. Additionally, we showed that sentence embeddings generated by \pmind exhibit cross-linguality.
    \item[$\bullet$] Showed the efficacy of multilingual transfer learning on \pmindns, evidencing that fine-tuning the model with Persian data notably enhances its performance on the corresponding English task.
\end{itemize}

The rest of this paper is organized as follows: 
\cref{sec:related_work} reviews open-source LLMs and parameter efficient fine-tuning methods. \cref{sec:pmind} details our training approach, while \cref{sec:evaluations} compares \pmindns's performance across various tasks with other competitors. In \cref{sec:carbon_footprint}, we discuss the carbon footprint associated with training \pmind. Finally, \cref{sec:conclusion} concludes the paper.

\section{Related Work}
\label{sec:related_work}
\subsection{Open LLMs}
While GPT-3.5-turbo demonstrated excellent proficiency in natural language generation, the LLaMa model~\citep{touvron2023llama} was the first LLM to claim achievement of similar performance in various English tasks.
LLaMa family of models are one of the most popular open foundation large language models, ranging in scale from 7B to 65B parameters.
Smaller LLaMa models are trained on 1T tokens, while larger ones are trained on 1.4T tokens.
In both cases, training data is predominantly consist of English text and code.
Only 4.5\% of their dataset is multilingual, including either Latin or Cyrillic scripts.
The Mosaic Pretrained Transformers (MPT) model~\citep{mosaic2023mpt} has 7B parameters and is trained on 1T tokens of English text and code.
This model increases the context length of inputs from 2k to 65k in its storyteller model.
LLaMa2~\citep{touvron2023llama2} models are an updated version of LLaMa models, comprising a collection of LLMs with 7B, 13B, and 70B parameters.
These models are pretrained on a larger and higher quality dataset compared to the first version of LLaMa models.
Their pretraining dataset primarily consists of English text and code, with less than 2\% of the text in other languages.

Falcon~\citep{tii2023falcom} models are a set of LLMs with 1.3B, 7.5B, 40B, and (closed-source) 180B parameters, trained on the RefinedWeb dataset~\citep{penedo2023refinedweb}
--- a curated, high-quality web-based dataset.
Although the RefinedWeb dataset is multilingual and includes many languages, such as Persian, open-source Falcon models are trained on European languages, particularly English.
Yi~\citep{yi202301} models are a collection of Chinese-English LLMs with 6B and 34B parameters.
These models are trained on a 3T token dataset of English and Chinese and outperform previous models on English and, especially, Chinese benchmarks.

While newer models like Falcon and Yi demonstrate better performance on English benchmarks, we decided to use Llama2 model due to training on more multilingual datasets.
We specifically opted to fine-tune the LLaMa2-7B-chat variant because loading it with the \texttt{fp16} data type requires only 14GB of GPU memory, making it easily loadable on a consumer GPU with 24GB of memory.
Loading our model on a single GPU allows us to avoid the overhead of model parallelism.
Furthermore, in a multi-GPU setup, this approach leads to faster training by leveraging data parallelism.

\begin{table*}[t]
    \centering
    \begin{tabular}{l l l l r}
        \toprule
        \multirow{2}{4em}{} & \multicolumn{2}{c}{Fine-Tuning} & \multicolumn{2}{c}{Instruction-Tuning} \\
        \cmidrule(lr){2-3} \cmidrule(lr){4-5}
        & Dataset & Num. Tokens & Dataset & Num. Intructions \\
        \cmidrule(lr){2-3} \cmidrule(lr){4-5}
        \textbf{Step 1} & Persian Wikipedia & 222M & -- & -- \\        \midrule
        \multirow{5}{4em}{\textbf{Step 2}} & \multirow{5}{10em}{CC100} & \multirow{5}{7em}{1.147B} & Alpaca (Fa) & 52,000 \\
        & & & TED2020 & 50,000 \\
        & & & ParsQuad & 5,000 \\
        & & & PersianQA & 900 \\
        \cmidrule{4-5}
        & & & Total & 107,900 \\
        \midrule
        \multirow{11}{4em}{\textbf{Step 3}} & \multirow{11}{10em}{CC100} & \multirow{11}{7em}{600M} & TED2020 & 300,000 \\
        & & & MMLU auxiliary (Fa) & 100,000 \\
        & & & CoT (Fa) & 74,000 \\
        & & & PN Summary & 20,000 \\
        & & & ParsQuad & 5,000 \\
        & & & Internal dataset & 1,241 \\
        & & & ParsiNLU Mutiple-Choice QA & 1,200 \\
        & & & PersianQA & 900 \\
        & & & ParsiNLU (Reading Comprehension) & 600 \\
        \cmidrule{4-5}
        & & & Total & 502,941 \\
        \bottomrule
    \end{tabular}
    \caption{Training and instruction tuning datasets for each step. Instruction dataset names with (Fa) indicate that they are Persian machine-translated datasets.}
    \label{tab:TrainingData}
\end{table*}

\subsection{Parameter Efficient Fine-Tuning}
While fine-tuning LLMs with billions of parameters can be an expensive task, \textit{parameter-efficient fine-tuning} (PEFT) techniques aim to reduce training costs by fine-tuning a small number of parameters.
Adapter tuning~\citep{houlsby2019parameter} was one of the first PEFT techniques.
In adapter tuning, small adapter layers are inserted after the multihead attention and feed forward layers of each transformer block, with only these layers undergoing training.
Adamix~\citep{wang-etal-2022-adamix} suggested leveraging adapters in a mixture-of-expert fashion, while SparseAdapters~\citep{he-etal-2022-sparseadapter} acknowledges the redundancy in many adapter parameters.
By pruning these redundant parameters during initialization and subsequently fine-tuning, SparseAdapter achieves better results.

Unlike adapter-based approaches, which often involve fine-tuning large models with the addition of small trainable parameters, a set of methods proposes a different approach --– fine-tuning only a small subset of the existing model.
BitFit~\citep{ben-zaken-etal-2022-bitfit} proposes fine-tuning only the biases of the model.
FishMask~\citep{sung2021training} recommends selecting parameters with the highest Fisher information value for training.
While Freeze and Reconfigure (FAR)~\citep{vucetic2022efficient} suggests freezing less important columns and only training the crucial ones.

\textit{Low-rank adaptation} (LoRA)~\citep{hu2022lora} approaches represent another category of PEFT methods.
LoRA introduces the concept of freezing the pretrained model and adding small, tunable weights into specific layers.
These weights take the form of the rank decomposition matrices derived from the pretrained model weights.
We refer interested readers to \citep{lialin2023scaling} which provides a comprehensive survey of PEFT approches.

Given the small number of trainable parameters and its great performance across models of various sizes (ranging from 125M to 175B), we choose to fine-tune our model utilizing the LoRA technique.
Moreover, the Bacterian-X~\citep{li2023bactrian} models are a notable example of the efficacy of LoRA in fine-tuning LLMs for learning an additional language.

\section{\pmind Model}
\label{sec:pmind}
\subsection{Bilingual Tokenizer and Expansion of Embeddings}
The LLaMa2's tokenizer has 32,000 tokens, including only 55 Arabic and Persian tokens.
It specifically covers Persian letters and does not include any additional Persian subwords.
In this situation, fine-tuning the model with a Persian corpus can take a long time because every word is tokenized into letters.
Therefore, we decide to train a new Persian tokenizer.
Augmenting the LLaMa2's tokenizer with Persian subwords enable fine-tuning our model on more extensive corpora within the constraints of the same computational budget.

Hence, our approach involve training a byte-pair encoding (BPE)~\citep{sennrich-etal-2016-neural} tokenizer with 10,000 tokens on a 1GB Persian Wikipedia corpus. 
Furthermore, we enhance the LLaMa2 tokenizer by incorporating Persian subwords.
The combined Persian-English tokenizer has 41,510 tokens.
Although training a larger BPE tokenizer can capture more named entity tokens, it would increase the input and output embeddings' size, resulting in more trainable parameters, which requires more computational resources.
Furthermore, we expand the input and output embeddings by the size of our tokenizer.
The newly added embeddings are randomly initialized within the space of LLaMa2's embeddings.

\subsection{Training Details}

\begin{figure*}[ht]
  \includegraphics[width=0.95\textwidth]{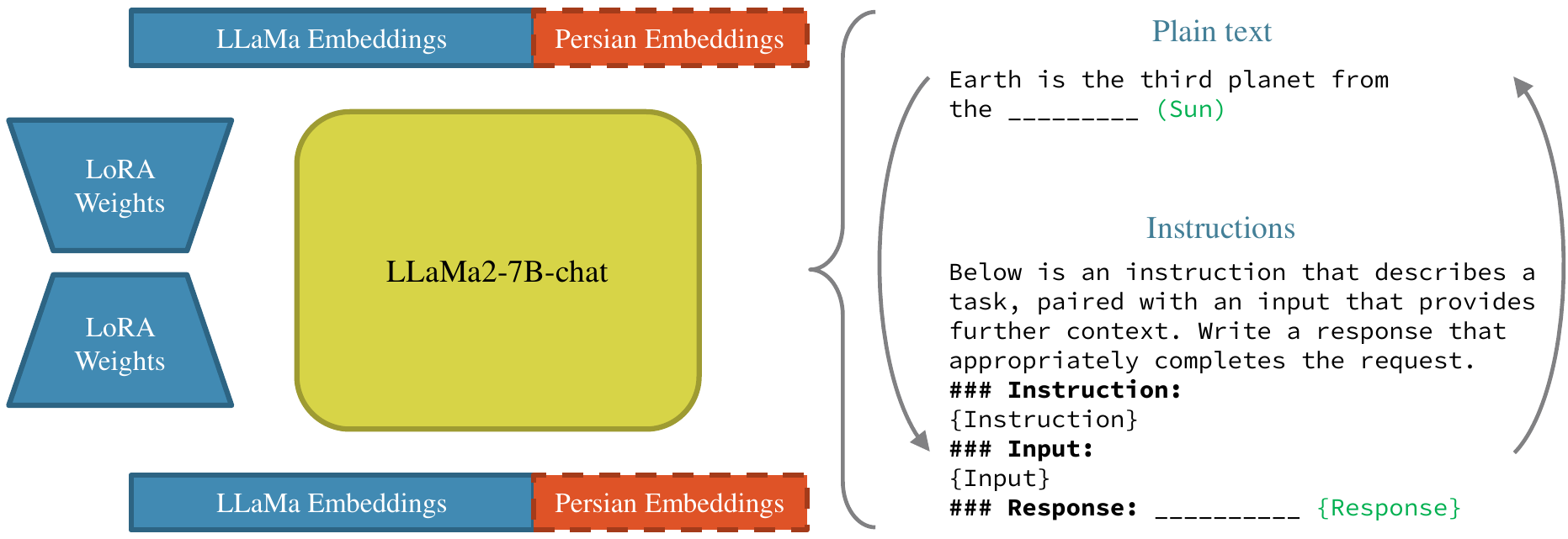}
  \caption{Our Approach for Training \pmindns: We expand LLaMa2's tokenizer and embeddings with 10,000 Persian subwords. Next, we employ the LoRA technique to reduce memory usage during training, with trainable components including input and output embeddings, as well as LoRA weights. The model is trained on Persian plain text and instructions iteratively, evaluating at each step. The training objective is focused on causal language modeling, wherein the model predicts the next token based on previously observed ones.}
  \label{fig:overall_trainig}
\end{figure*}

Our model training approach is focused on optimizing training on constrained computational budget.
To achieve this, we leverage the 7B-chat variant of LLaMa2 models.
We avoid model parallelism due to significant overhead in the initial release of \pmind and will consider it for the future releases.
Instead, we embrace data parallelism to reduce training time.
Consequently, the LoRA technique is implemented by incorporating LoRA weights across all layers of the transformer architecture, utilizing a LoRA rank of 8 and a dropout rate of 0.05.
Therefore, our training parameters encompass LoRA weights, input, and output embeddings.

Our training methodology comprises two key phases.
First, we fine-tune the embeddings and LoRA weights of the model through training on a plain Persian text corpus.
Subsequently, we perform supervised fine-tuning (SFT) of the model on instruction datasets.
During both phases, our training objective is casual language modeling.
To ensure a thorough evaluation, we divide the training into three distinct steps.

\textbf{Step 1: } In the first step, we fine-tune our model for two epochs on the Persian Wikipedia corpus, which contains 111 million tokens. 
During this phase, we intentionally skip fine-tuning on instructions to explore the model's capabilities. 
Despite the less-than-ideal perplexity, the model still manage to produce acceptable results.

\textbf{Step 2: } In the second step, we fine-tune our model on a 10GB subset of CC100~\citep{conneau-etal-2020-unsupervised} dataset, which includes 1.147 billion tokens. 
In this phase, we fine-tun our model on PersianQA~\citep{PersianQA} and ParsQuad~\citep{Parsquad} datasets to improve its performance in the question-answering task. 
Additionally, we perform fine-tuning on a segment of the English-Persian TED2020~\citep{reimers-2020-multilingual-sentence-bert} parallel dataset for better translations, and fine-tune on machine-translated Alpaca~\citep{alpaca} dataset, a comprehensive instructions dataset covering various tasks and questions.

\textbf{Step 3: } In the third step, we fine-tune our model on an additional 5GB subset of the CC100 dataset, encompassing 600 million tokens. 
This phase prioritize fine-tuning on a variety of instructions datasets.
Notably, to enhance our model's translation capabilities, we fine-tune on the entire English-Persian TED2020 parallel dataset.
PersianQA and ParsQuad are once again employed to augment model's capabilities in the question-answering task.
To improve proficiency in multiple-choice QA and reading comprehension tasks, we fine-tune our model on ParsiNLU multiple-choice QA~\citep{ParsiNLU} , ParsiNLU reading comprehension~\citep{ParsiNLU}, and the machine-translated auxiliary dataset of MMLU~\citep{hendryckstest2021}.
Additionally, 20,000 instructions from the PN summary dataset~\citep{pnSummary} are incorporated to strengthen our model's summarization capabilities. 
Natural language reasoning capabilities are refined by fine-tuning on the machine-translated ORCA's Chain of Thought dataset~\citep{mukherjee2023orca}. 
Furthermore, a few manually created instructions related to Persian food recipes and proverbs are introduced to diversify the model's understanding of natural language contexts.

The model trained in this step exhibits proficiency in generating Persian text, with a perplexity score superior to that of the second-step model.
Additionally, there is a slight improvement in the perplexity score of the model when applied to the English text. Notably, fine-tuning on the ORCA's Chain of Thought dataset contributed to enhancing the model's ability to analyze and solve problems systematically and step by step.

\Cref{fig:overall_trainig} provides an overview of our training approach, while \Cref{tab:TrainingData} presents the datasets used for training, along with their respective token counts. 
Additionally, it provides information on the datasets used for instruction tuning in each step, along with the corresponding number of instructions.

\begin{table}
    \centering
    \begin{tabular}{lcccc}
        \toprule
        & \multicolumn{2}{c}{After Fine-Tuning} & \multicolumn{2}{c}{After Instruction-Tuning} \\
        \cmidrule(lr){2-3} \cmidrule(lr){4-5}
        & English & Persian & English & Persian \\
        \midrule
        Step 1 & 15.10 & 34.09 & - & - \\
        Step 2 & 24.18 & 26.56 & 14.89 & 16.95 \\
        Step 3 & 29.04 & 20.35 & 12.30 & 14.55 \\
        \bottomrule
    \end{tabular}
    \caption{
    Perplexity scores for Persian and English evaluation datasets across training steps after fine-tuning and instruction-tuning.
    Note that LLaMa2-7B-chat achieves a perplexity score of 8.82 on the English evaluation dataset.}
    \label{tab:perplexity}
\end{table}

\Cref{tab:perplexity} displays the perplexity scores observed in Persian and English datasets after each step of our model's training, before and after SFT. 
The English perplexity score was computed on an English corpus comprising 24 articles from The~New~York~Times\footnote{\url{https://www.nytimes.com/}} and The~New~Yorker\footnote{\url{https://www.newyorker.com/}}, covering various subjects and containing 54,000 tokens.
For the Persian perplexity score, we selected nine articles from Tarjomaan\footnote{\url{https://tarjomaan.com/}} and Faradars\footnote{\url{https://blog.faradars.org/}}, spanning various domains and containing 58,000 tokens.
For both English and Persian datasets, we selected recently published articles to ensure that our model was not trained on them.

The results indicate that SFT has a significant impact on perplexity scores for both Persian and English languages. 
In the second step, training our model on more than 1 billion plain Persian text only resulted in an 8-point improvement in perplexity score. 
However, performing SFT on 170,000 instructions led to a 10-point improvement.

Notably, perplexity scores show that training on plain Persian text could lead to catastrophic forgetting of the model's English knowledge. 
However, performing SFT on Persian-English parallel dataset could help mitigate this effect.

\section{Evaluations}
\label{sec:evaluations}
\subsection{Multiple-Choice QA and Reading Comprehension}
We evaluated our model using the test subset of ParsiNLU multiple-choice QA and Belebele multiple-choice reading comprehension dataset~\citep{bandarkar2023belebele}. 
The ParsiNLU dataset consists of Persian questions with multiple candidates, where one of them is the correct answer.
Notably, there is no context paragraph for these questions.
In contrast, the Belebele dataset presents questions extracted from provided paragraphs, with four candidate answers, with one being the correct response.

\begin{table}
    \centering
    \resizebox{\columnwidth}{!}{
    \begin{tabular}{llcccc}
        \toprule
        & Model & Lit. & Com. Know. & Math & All \\
        \midrule
        Zero-shot & \pmind & 35.7 & \underline{43.1} & 29.4 & 36.1 \\
        \midrule
        \multirow{7}{*}{Fine-tuned} & mT5(l)-fa & 32.6 & 27.1 & \textbf{38.9} & 32.9 \\
        & mT5(xl)-fa & {33.7} & 27.7 & \textbf{38.9} & 33.4 \\
        & mT5(l)-en & 27.4 & 33.1 & 25.4 & 28.6 \\
        & mT5(xl)-en & 28.3 & 38.6 & 22.0 & 29.6 \\
        & mT5(l)-fa\&en & 30.6 & 28.9 & 38.6 & 32.7 \\
        & mT5(xl)-fa\&en & \underline{38.0} & 33.7 & 38.0 & \underline{36.6} \\
        \cmidrule{2-6}
        & \pmind & \textbf{39.7} & \textbf{45.4} & \underline{38.8} & \textbf{41.3} \\
        \bottomrule
    \end{tabular}
    }
    \caption{Comparison of \pmindns's performance in ParsiNLU multiple-choice QA dataset across literature, common knowledge, and math \& logic Categories.}
    \label{tab:mcqa}
\end{table}

\Cref{tab:mcqa} shows evaluation results on the ParsiNLU multiple-choice QA dataset, categorized into \textit{literature}, \textit{common knowledge}, and \textit{math \& logic} question types.
The results also involve comparing our model with the large and x-large variants of mT5 models~\citep{xue-etal-2021-mt5}, each trained on Persian, English, or a mixture of both training datasets.
The evaluation encompasses both the \pmind model and its fine-tuned version.
The \pmind was fine-tuned for a single epoch on the training dataset, whereas mT5 model was fine-tuned for 20k steps on the training set for at least 15 epoch~\citep{ParsiNLU}.

As can be seen, \pmind achieved comparable results to the mT5 x-large model trained on both Persian and English datasets.
Remarkably, the fine-tuned \pmind model outperformed the best-performing model by almost 5\%.
While it demonstrates relatively weaker performance on \textit{math \& logic} questions, it excels in the \textit{common knowledge} category, outperforming other models in this specific domain.

\begin{table}
    \centering
    \begin{tabular}{lcc}
        \toprule
        Model & English & Persian \\
        \midrule
        GPT-3.5-tubro & \textbf{87.7} & 61.8 \\
        LLaMa2-7b-chat & 43.9 & 25.6 \\
        XLM-V large & 77.8 & \underline{70.8} \\
        \midrule
        \pmind & \underline{86.4} & \textbf{73.9} \\
        \bottomrule
    \end{tabular}
    \caption{Accuracy comparison of \pmind with the original LLaMa2-7B-chat, the multilingual masked language model XLM-V, and GPT-3.5-turbo on Belebele benchmark.}
    \label{tab:Belebele}
\end{table}

We evaluated \pmind on both the Persian and English subsets of the Belebele dataset. 
In \Cref{tab:Belebele}, we compared the results of our model with recently released multilingual masked language model XLM-V~\citep{liang-etal-2023-xlm}, the original LLaMa2-7B-chat, and GPT-3.5-turbo.
Results indicate that \pmind enhanced the performance of LLaMa2-7B-chat by 43\% and 48\% on English and Persian subsets, respectively.
Additionally, \pmind outperformed the XLM-V model.
Although its performance in English was 1\% lower  than that of GPT-3.5-turbo, it surpassed GPT-3.5-turbo by 12\% in the Persian subset.

Despite \pmind was fine-tuned exclusively on Persian multiple-choice QA datasets in step 3, its performance in English was also improved.
This observation highlights the potential for multilingual transfer learning for LLMs.

\begin{table*}[ht]
    \centering
    \begin{tabular}{ m{15em}  m{4em}  m{4em}  m{4em}  m{4em} }
        \toprule
        & \multicolumn{2}{c}{Fa$\rightarrow$En} & \multicolumn{2}{c}{En$\rightarrow$Fa} \\
        \cmidrule(lr){2-3}
        \cmidrule(lr){4-5}
        Model & BLEU & \textsc{Comet} & BLEU & \textsc{Comet} \\ \cmidrule{1-5}
        GPT-3.5-turbo & \underline{31.7} & \underline{87.37} & \underline{18.2} & \underline{84.66} \\
        mT5(l)-ParsiNLU (fine-tuned) & 23.3 & 82.39 & 15.8 & 83.77 \\
        NLLB-MoE & \textbf{40.9} & \textbf{88.70} & \textbf{24.9} & \textbf{87.75} \\
        \midrule
        \pmind (zero-shot) & 13.1 & 73.51 & 12.7 & 78.75 \\
        \pmind (2-shot) & 15.4 & 79.68 & 11.0 & 77.25 \\
        \pmind (4-shot) & 15.4 & 79.92 & 10.2 & 77.73 \\
        \pmind (fine-tuned)  & 25.7 & 83.61 & 15.4 & 79.44 \\
        \bottomrule
    \end{tabular}
    \caption{BLEU and \textsc{Comet} scores of \pmind in various setups, compared against mT5-large, GPT-3.5-turbo, and NLLB-MoE for for Fa$\rightarrow$En and En$\rightarrow$Fa translation directions. Note that \pmind~(fine-tuned) was fine-tuned only on 5,000 parallel sentences, while mT5(l)-ParsiNLU (fine-tuned) was fine-tuned on 200,000 Persian-English parallel sentences.}
    \label{tab:Translation}
\end{table*}

\subsection{Translation}
We evaluated \pmind on the \textsc{Flores}-200 dataset~\citep{nllb2022} for Fa$\rightarrow$En and En$\rightarrow$Fa translation directions.
Model evaluations were performed using both BLEU~\citep{post-2018-call} and \textsc{Comet}~\citep{rei-etal-2022-comet} scores, and results are presented in \Cref{tab:Translation}.
Our model was compared in various setups, including zero-shot, 2-shot, and 4-shot.
Additionally, we fine-tuned \pmind on 5,000 parallel data instances from WikiMatrix~\citep{schwenk-etal-2021-wikimatrix} and compared its performance with other models.
The ParsiNLU mT5-large model was fine-tuned for 200k steps on Persian-English parallel datasets~\citep{ParsiNLU}.
Furthermore, we benchmarked our model against GPT-3.5-turbo and NLLB-MoE~\citep{nllb2022}, a multilingual translation model with 54B parameters, supporting 200 languages.

Results show that while \pmindns's Fa$\rightarrow$En translation improved in a few-shot setup, the translation quality of En$\rightarrow$Fa declined.
This observation suggests that in-context learning is not an effective approach for the translation task.
Subsequently, we fine-tuned \pmind on 5,000 translation instructions. 
The results demonstrate that fine-tuned \pmind could generate results on a par with mT5-large model which was fine-tuned on a 200,000 Persian-English parallel dataset. 
However, fine-tuned \pmind still generated weaker results compared to GPT-3.5-turbo.

In our comparison, we additionally evaluated LLM results against the NLLB-MoE translation model.
Despite a substantial gap between LLM results and NLLB-MoE in terms of BLEU score, the gap is much smaller in \textsc{Comet} score.
This conveys the fact that while LLM translations have very similar meanings to the reference translations, there is not a high overlap between n-gram of model translations and reference translations.

\subsection{Semantic Textual Similarity}

To assess the quality of sentence embeddings generated by our model, we conducted evaluations on Semantic Textual Similarity (STS) benchmarks~\citep{corley2005measuring}. 
Initially, we evaluated our model's sentence embeddings on both Persian and English STS datasets independently.
Subsequently, we evaluated the cross-lingual performance of semantically similar sentences in the Persian-English context.

For Persian, we utilized the FarSick dataset~\citep{ghasemi2021farsick}, providing sentence pairs with relatedness scores ranging from 1.0 to 5.0. 
For English, we employed MTEB's STS benchmark dataset~\citep{muennighoff-etal-2023-mteb}, structured similarly to the FarSick dataset. 
In our evaluation process, we generated embedding for each sentence in a sentence pair separately and computed their relatedness using cosine similarity. 
Subsequently, we compared the model's similarity scores with the gold scores using the Spearman correlation metric.

We generated sentence embedding with our model using the \textit{AnglE approach}~\citep{li2023angle}, utilizing the embedding of the padding token in the following prompt template: \texttt{Summarize sentence "{text}" in one word:} 
Our results were then compared to other multilingual foundation language models, including mBERT\citep{devlin-etal-2019-bert}, ~ParsBERT\citep{ParsBERT}, and XLM-RoBERTa~\citep{conneau-etal-2020-unsupervised}. 
The computation of sentence embedding from these models involved various approaches, such as utilizing the \texttt{[CLS]} token's embedding, mean pooling, and the AnglE-BERT approach.
Subsequently, we compared the best results from other models with the results obtained from our model.
Additionally, we evaluated our model's results in comparison to LaBSE~\citep{feng-etal-2022-language} and LASER3~\citep{heffernan-etal-2022-bitext} models, which are commonly employed for bitext mining purposes.

\begin{table}
    \centering
    \resizebox{\columnwidth}{!}{
    \begin{tabular}{lccc}
        \toprule
        & FarSick & STS &  \\
        \cmidrule(lr){2-2}
        \cmidrule(lr){3-3}
        Model & Spearman Corr. & Spearman Corr. & Avg. \\
        \cmidrule{1-4}
        LaBSE & \textbf{66.79} & 72.25 & \underline{69.52} \\
        LASER3 & 60.57 & 69.77 & 65.17 \\
        LLaMa2-7B-chat & 54.24 & \underline{73.65} & 63.94 \\
        mBERT & 52.09 & 50.97 & 51.53 \\
        ParsBERT & 54.73 & 51.40 & 53.06 \\
        XLM-RoBERTa & 49.45 & 34.49 & 41.97 \\
        \midrule
        \pmind & \underline{63.76} & \textbf{75.73} & \textbf{69.74} \\
        \bottomrule
    \end{tabular}
    }
    \caption{Comparison of \pmindns's semantic similarity scores on FarSick and STS-benchmark datasets with other foundation language and bitext mining models.}
    \label{tab:sts}
\end{table}

In \Cref{tab:sts}, we present the results of semantic similarity scores for \pmind and other models. 
The findings demonstrate that \pmind achieved the highest Spearman correlation score among all foundation language models, both in English and Persian. 
It showcased improvement over LLaMa2-7B-chat correlation scores by 9\% in Persian and 2\% in English. 
Furthermore, \pmind outperformed LASER3 sentence embeddings and achieved comparable average results with LaBSE.

To assess the cross-linguality of our model's sentence embedding, we utilized the Persian-English subset of the \textsc{Flores}-200 dataset. 
For evaluating the semantic similarity of parallel English-Persian sentences, we computed the cosine similarity of Persian and English sentence embedding and then averaged across all pairs.
\Cref{tab:CrossLingual} compares \pmind against other foundation language and bitext mining models. 
While \pmind did not reach the average cosine similarity score of LaBSE, a model specialized for this task, it still achieved a notable 72\%, surpassing the original model by 14\% and emerging as the top-performing language model among others.

\begin{table}
    \centering
    \begin{tabular}{lccc}
        \toprule
        Model & Avg. Cosine Similarity  \\
        \midrule
        LaBSE & \textbf{86.86} \\
        LASER3 & \underline{74.49} \\
        LLaMa2-7B-chat & 58.68 \\
        mBERT & 64.06 \\
        ParsBERT & 48.29 \\
        \midrule
        \pmind & 72.36 \\
        \bottomrule
    \end{tabular}
    \caption{Cross-lingual semantic similarity comparison of \pmind against other foundation language and bitext mining models on the English-Persian subset of the \textsc{Flores} Dataset.}
    \label{tab:CrossLingual}
\end{table}

\section{Carbon Footprint}
\label{sec:carbon_footprint}
\pmind was trained on four NVIDIA~RTX~3090 GPUs. 
Training on Persian plain text required 9~days, while performing the SFT on instructions took 1~day. 
Consequently, the training of \pmind consumed a total of 960 GPU-hours. 
The power consumption of each NVIDIA~RTX~3090 GPU is 350W. 
Additionally, considering a Power Usage Effectiveness (PUE) of 1.4 for our servers due to suboptimal equipment efficiency, the overall power consumption for training \pmindns, based on the formula \citep{wu2022sustainable}, was 470kWh.

To calculate carbon emissions, we need to estimate the CO$_2$ emission intensity, which is linked to the location of our data center. 
We utilized global electricity data~\citep{CarbonIntensity} on Iran's carbon intensity of electricity, which measured 494 grams of CO$_2$ equivalents emitted per kilowatt-hour of electricity. 
Therefore, the training of \pmind resulted in emitting 232.38 kCO$_2$eq.

\section{Conclusion}
\label{sec:conclusion}
In this paper, we introduced \pmindns, a large language model built upon LLaMa2-7B-chat as the foundation model which incorporated additional 10,000 Persian subwords and trained on an extensive dataset of almost 2 billion Persian tokens. 
By employing LoRA in our training, our aim was to achieve cost-effective training.
We demonstrated that employing the LoRA technique and conducting SFT on English-Persian parallel datasets allows us to reduce catastrophic forgetting of English knowledge despite our model being trained on a large collection of Persian datasets.
Our findings indicate that \pmind achieved comparable results in reading comprehension multiple-choice QA datasets with GPT-3.5-turbo. 
By fine-tuning with 5,000 parallel sentences, it surpassed the performance of the mT5-large model, which was fine-tuned with 200,000 samples, in translation tasks. 
Additionally, we demonstrated that \pmind produced highly effective sentence embedding for both English and Persian sentences. 
As a language model, it outperformed previous masked language models, showcasing its efficacy in various natural language processing tasks.

\printbibliography

@article{bubeck2023sparks,
  title={{Sparks of artificial general intelligence: Early experiments with GPT-4}},
  author={Bubeck, S{\'e}bastien and Chandrasekaran, Varun and Eldan, Ronen and Gehrke, Johannes and Horvitz, Eric and Kamar, Ece and Lee, Peter and Lee, Yin Tat and Li, Yuanzhi and Lundberg, Scott and others},
  journal={arXiv preprint arXiv:2303.12712},
  year={2023}
}

@article{vaswani2017attention,
  title={{Attention is all you need}},
  author={Vaswani, Ashish and Shazeer, Noam and Parmar, Niki and Uszkoreit, Jakob and Jones, Llion and Gomez, Aidan N and Kaiser, {\L}ukasz and Polosukhin, Illia},
  journal={Advances in neural information processing systems},
  year={2017}
}

@article{zhao2023survey,
  title={{A survey of large language models}},
  author={Zhao, Wayne Xin and Zhou, Kun and Li, Junyi and Tang, Tianyi and Wang, Xiaolei and Hou, Yupeng and Min, Yingqian and Zhang, Beichen and Zhang, Junjie and Dong, Zican and others},
  journal={arXiv preprint arXiv:2303.18223},
  year={2023}
}

@article{wu2023visual,
  title={{Visual chatgpt: Talking, drawing and editing with visual foundation models}},
  author={Wu, Chenfei and Yin, Shengming and Qi, Weizhen and Wang, Xiaodong and Tang, Zecheng and Duan, Nan},
  journal={arXiv preprint arXiv:2303.04671},
  year={2023}
}

@misc{openai2023chatgpt,
      title={{ChatGPT}}, 
      author={OpenAI},
      year={2022},
      Note={\url{https://openai.com/blog/chatgpt} [Accessed: 01-07-2024]}
}

@misc{openai2023gpt4,
      title={GPT-4 Technical Report}, 
      author={OpenAI and : and Josh Achiam and Steven Adler and Sandhini Agarwal and Lama Ahmad and Ilge Akkaya and Florencia Leoni Aleman and Diogo Almeida and Janko Altenschmidt and Sam Altman and Shyamal Anadkat and Red Avila and Igor Babuschkin and Suchir Balaji and Valerie Balcom and Paul Baltescu and Haiming Bao and Mo Bavarian and Jeff Belgum and Irwan Bello and Jake Berdine and Gabriel Bernadett-Shapiro and Christopher Berner and Lenny Bogdonoff and Oleg Boiko and Madelaine Boyd and Anna-Luisa Brakman and Greg Brockman and Tim Brooks and Miles Brundage and Kevin Button and Trevor Cai and Rosie Campbell and Andrew Cann and Brittany Carey and Chelsea Carlson and Rory Carmichael and Brooke Chan and Che Chang and Fotis Chantzis and Derek Chen and Sully Chen and Ruby Chen and Jason Chen and Mark Chen and Ben Chess and Chester Cho and Casey Chu and Hyung Won Chung and Dave Cummings and Jeremiah Currier and Yunxing Dai and Cory Decareaux and Thomas Degry and Noah Deutsch and Damien Deville and Arka Dhar and David Dohan and Steve Dowling and Sheila Dunning and Adrien Ecoffet and Atty Eleti and Tyna Eloundou and David Farhi and Liam Fedus and Niko Felix and Simón Posada Fishman and Juston Forte and Isabella Fulford and Leo Gao and Elie Georges and Christian Gibson and Vik Goel and Tarun Gogineni and Gabriel Goh and Rapha Gontijo-Lopes and Jonathan Gordon and Morgan Grafstein and Scott Gray and Ryan Greene and Joshua Gross and Shixiang Shane Gu and Yufei Guo and Chris Hallacy and Jesse Han and Jeff Harris and Yuchen He and Mike Heaton and Johannes Heidecke and Chris Hesse and Alan Hickey and Wade Hickey and Peter Hoeschele and Brandon Houghton and Kenny Hsu and Shengli Hu and Xin Hu and Joost Huizinga and Shantanu Jain and Shawn Jain and Joanne Jang and Angela Jiang and Roger Jiang and Haozhun Jin and Denny Jin and Shino Jomoto and Billie Jonn and Heewoo Jun and Tomer Kaftan and Łukasz Kaiser and Ali Kamali and Ingmar Kanitscheider and Nitish Shirish Keskar and Tabarak Khan and Logan Kilpatrick and Jong Wook Kim and Christina Kim and Yongjik Kim and Hendrik Kirchner and Jamie Kiros and Matt Knight and Daniel Kokotajlo and Łukasz Kondraciuk and Andrew Kondrich and Aris Konstantinidis and Kyle Kosic and Gretchen Krueger and Vishal Kuo and Michael Lampe and Ikai Lan and Teddy Lee and Jan Leike and Jade Leung and Daniel Levy and Chak Ming Li and Rachel Lim and Molly Lin and Stephanie Lin and Mateusz Litwin and Theresa Lopez and Ryan Lowe and Patricia Lue and Anna Makanju and Kim Malfacini and Sam Manning and Todor Markov and Yaniv Markovski and Bianca Martin and Katie Mayer and Andrew Mayne and Bob McGrew and Scott Mayer McKinney and Christine McLeavey and Paul McMillan and Jake McNeil and David Medina and Aalok Mehta and Jacob Menick and Luke Metz and Andrey Mishchenko and Pamela Mishkin and Vinnie Monaco and Evan Morikawa and Daniel Mossing and Tong Mu and Mira Murati and Oleg Murk and David Mély and Ashvin Nair and Reiichiro Nakano and Rajeev Nayak and Arvind Neelakantan and Richard Ngo and Hyeonwoo Noh and Long Ouyang and Cullen O'Keefe and Jakub Pachocki and Alex Paino and Joe Palermo and Ashley Pantuliano and Giambattista Parascandolo and Joel Parish and Emy Parparita and Alex Passos and Mikhail Pavlov and Andrew Peng and Adam Perelman and Filipe de Avila Belbute Peres and Michael Petrov and Henrique Ponde de Oliveira Pinto and Michael and Pokorny and Michelle Pokrass and Vitchyr Pong and Tolly Powell and Alethea Power and Boris Power and Elizabeth Proehl and Raul Puri and Alec Radford and Jack Rae and Aditya Ramesh and Cameron Raymond and Francis Real and Kendra Rimbach and Carl Ross and Bob Rotsted and Henri Roussez and Nick Ryder and Mario Saltarelli and Ted Sanders and Shibani Santurkar and Girish Sastry and Heather Schmidt and David Schnurr and John Schulman and Daniel Selsam and Kyla Sheppard and Toki Sherbakov and Jessica Shieh and Sarah Shoker and Pranav Shyam and Szymon Sidor and Eric Sigler and Maddie Simens and Jordan Sitkin and Katarina Slama and Ian Sohl and Benjamin Sokolowsky and Yang Song and Natalie Staudacher and Felipe Petroski Such and Natalie Summers and Ilya Sutskever and Jie Tang and Nikolas Tezak and Madeleine Thompson and Phil Tillet and Amin Tootoonchian and Elizabeth Tseng and Preston Tuggle and Nick Turley and Jerry Tworek and Juan Felipe Cerón Uribe and Andrea Vallone and Arun Vijayvergiya and Chelsea Voss and Carroll Wainwright and Justin Jay Wang and Alvin Wang and Ben Wang and Jonathan Ward and Jason Wei and CJ Weinmann and Akila Welihinda and Peter Welinder and Jiayi Weng and Lilian Weng and Matt Wiethoff and Dave Willner and Clemens Winter and Samuel Wolrich and Hannah Wong and Lauren Workman and Sherwin Wu and Jeff Wu and Michael Wu and Kai Xiao and Tao Xu and Sarah Yoo and Kevin Yu and Qiming Yuan and Wojciech Zaremba and Rowan Zellers and Chong Zhang and Marvin Zhang and Shengjia Zhao and Tianhao Zheng and Juntang Zhuang and William Zhuk and Barret Zoph},
      year={2023},
      eprint={2303.08774},
      archivePrefix={arXiv},
      primaryClass={cs.CL}
}

@article{anil2023palm,
  title={{Palm 2 technical report}},
  author={Anil, Rohan and Dai, Andrew M and Firat, Orhan and Johnson, Melvin and Lepikhin, Dmitry and Passos, Alexandre and Shakeri, Siamak and Taropa, Emanuel and Bailey, Paige and Chen, Zhifeng and others},
  journal={arXiv preprint arXiv:2305.10403},
  year={2023}
}

@misc{anthropic2023claude,
      title={{Claude}}, 
      author={Anthropic},
      year={2023},
      Note={\url{https://claude.ai} [Accessed: 01-07-2024]}
}

@article{touvron2023llama2,
  title={{Llama 2: Open foundation and fine-tuned chat models}},
  author={Touvron, Hugo and Martin, Louis and Stone, Kevin and Albert, Peter and Almahairi, Amjad and Babaei, Yasmine and Bashlykov, Nikolay and Batra, Soumya and Bhargava, Prajjwal and Bhosale, Shruti and others},
  journal={arXiv preprint arXiv:2307.09288},
  year={2023}
}

@article{touvron2023llama,
  title={{Llama: Open and efficient foundation language models}},
  author={Touvron, Hugo and Lavril, Thibaut and Izacard, Gautier and Martinet, Xavier and Lachaux, Marie-Anne and Lacroix, Timoth{\'e}e and Rozi{\`e}re, Baptiste and Goyal, Naman and Hambro, Eric and Azhar, Faisal and others},
  journal={arXiv preprint arXiv:2302.13971},
  year={2023}
}

@misc{mosaic2023mpt,
      title={{MPT}}, 
      author={MosaicML},
      year={2023},
      Note="\url{https://www.mosaicml.com/blog/mpt-7b} [Accessed: 01-07-2024]"
}

@article{penedo2023refinedweb,
  title={{The RefinedWeb dataset for Falcon LLM: outperforming curated corpora with web data, and web data only}},
  author={Penedo, Guilherme and Malartic, Quentin and Hesslow, Daniel and Cojocaru, Ruxandra and Cappelli, Alessandro and Alobeidli, Hamza and Pannier, Baptiste and Almazrouei, Ebtesam and Launay, Julien},
  journal={arXiv preprint arXiv:2306.01116},
  year={2023}
}

@misc{tii2023falcom,
      title={Falcon}, 
      author={Tii},
      year={2023},
      Note={\url{https://falconllm.tii.ae/falcon.html} [Accessed: 01-07-2024]}
}

@misc{yi202301,
      title={01}, 
      author={Yi},
      year={2023},
      Note={\url{https://01.ai/} [Accessed: 01-07-2024]}
}

@inproceedings{houlsby2019parameter,
  title={{Parameter-efficient transfer learning for NLP}},
  author={Houlsby, Neil and Giurgiu, Andrei and Jastrzebski, Stanislaw and Morrone, Bruna and De Laroussilhe, Quentin and Gesmundo, Andrea and Attariyan, Mona and Gelly, Sylvain},
  booktitle={International Conference on Machine Learning},
  year={2019},
}

@inproceedings{wang-etal-2022-adamix,
    title = "{{AdaMix: Mixture-of-Adaptations for Parameter-efficient Model Tuning}}",
    author = "Wang, Yaqing  and
      Agarwal, Sahaj  and
      Mukherjee, Subhabrata  and
      Liu, Xiaodong  and
      Gao, Jing  and
      Awadallah, Ahmed Hassan  and
      Gao, Jianfeng",
    booktitle = "Proceedings of the 2022 Conference on Empirical Methods in Natural Language Processing",
    year = "2022",
}

@inproceedings{he-etal-2022-sparseadapter,
    title = "{{SparseAdapter: An Easy Approach for Improving the Parameter-Efficiency of Adapters}}",
    author = "He, Shwai  and
      Ding, Liang  and
      Dong, Daize  and
      Zhang, Jeremy  and
      Tao, Dacheng",
    booktitle = "Findings of the Association for Computational Linguistics",
    year = "2022",
}

@inproceedings{ben-zaken-etal-2022-bitfit,
    title = "{{BitFit: Simple Parameter-efficient Fine-tuning for Transformer-based Masked Language-models}}",
    author = "Ben Zaken, Elad  and
      Goldberg, Yoav  and
      Ravfogel, Shauli",
    booktitle = "Proceedings of the 60th Annual Meeting of the Association for Computational Linguistics",
    year = "2022",
}

@article{sung2021training,
  title={{Training neural networks with fixed sparse masks}},
  author={Sung, Yi-Lin and Nair, Varun and Raffel, Colin A},
  journal={Advances in Neural Information Processing Systems},
  year={2021}
}

@inproceedings{vucetic2022efficient,
  title={{Efficient fine-tuning of bert models on the edge}},
  author={Vucetic, Danilo and Tayaranian, Mohammadreza and Ziaeefard, Maryam and Clark, James J and Meyer, Brett H and Gross, Warren J},
  booktitle={2022 IEEE International Symposium on Circuits and Systems},
  year={2022},
}

@inproceedings{hu2022lora,
    title={{LoRA: Low-Rank Adaptation of Large Language Models}},
    author={Edward J Hu and Yelong Shen and Phillip Wallis and Zeyuan Allen-Zhu and Yuanzhi Li and Shean Wang and Lu Wang and Weizhu Chen},
    booktitle={International Conference on Learning Representations},
    year={2022},
}

@inproceedings{sennrich-etal-2016-neural,
    title = "{{Neural Machine Translation of Rare Words with Subword Units}}",
    author = "Sennrich, Rico  and
      Haddow, Barry  and
      Birch, Alexandra",
    booktitle = "Proceedings of the 54th Annual Meeting of the Association for Computational Linguistics",
    year = "2016",
}

@inproceedings{conneau-etal-2020-unsupervised,
    title = "{{Unsupervised Cross-lingual Representation Learning at Scale}}",
    author = "Conneau, Alexis  and
      Khandelwal, Kartikay  and
      Goyal, Naman  and
      Chaudhary, Vishrav  and
      Wenzek, Guillaume  and
      Guzm{\'a}n, Francisco  and
      Grave, Edouard  and
      Ott, Myle  and
      Zettlemoyer, Luke  and
      Stoyanov, Veselin",
    booktitle = "Proceedings of the 58th Annual Meeting of the Association for Computational Linguistics",
    year = "2020",
}

@inproceedings{reimers-2020-multilingual-sentence-bert,
    title = "{{Making Monolingual Sentence Embeddings Multilingual using Knowledge Distillation}}",
    author = "Reimers, Nils and Gurevych, Iryna",
    booktitle = "Proceedings of the 2020 Conference on Empirical Methods in Natural Language Processing",
    year = "2020",
}

@misc{alpaca,
  author = {Rohan Taori and Ishaan Gulrajani and Tianyi Zhang and Yann Dubois and Xuechen Li and Carlos Guestrin and Percy Liang and Tatsunori B. Hashimoto },
  title = {{Stanford Alpaca: An Instruction-following LLaMA Model}},
  year = {2023},
  publisher = {GitHub},
  journal = {GitHub repository},
  howpublished = {\url{https://github.com/tatsu-lab/stanford_alpaca}},
}

@misc{PersianQA,
  author          = {Ayoubi, Sajjad and Davoodeh, Mohammad Yasin},
  title           = {{PersianQA: a dataset for Persian Question Answering}},
  year            = 2021,
  publisher       = {GitHub},
  journal         = {GitHub repository},
  howpublished    = {\url{https://github.com/SajjjadAyobi/PersianQA}},
}

@article {Parsquad,
author = {Abadani, Negin and Mozafari, Jamshid and Fatemi,    Afsaneh and Nematbakhsh, Mohamadali and Kazemi, Arefeh},
title = {{ParSQuAD: Persian Question Answering Dataset based on Machine Translation of SQuAD 2.0}},
journal = {International Journal of Web Research},
year  = {2021},
}

@article{ParsiNLU,
    author = {Khashabi, Daniel and Cohan, Arman and Shakeri, Siamak and Hosseini, Pedram and Pezeshkpour, Pouya and Alikhani, Malihe and Aminnaseri, Moin and Bitaab, Marzieh and Brahman, Faeze and Ghazarian, Sarik and Gheini, Mozhdeh and Kabiri, Arman and Mahabagdi, Rabeeh Karimi and Memarrast, Omid and Mosallanezhad, Ahmadreza and Noury, Erfan and Raji, Shahab and Rasooli, Mohammad Sadegh and Sadeghi, Sepideh and Azer, Erfan Sadeqi and Samghabadi, Niloofar Safi and Shafaei, Mahsa and Sheybani, Saber and Tazarv, Ali and Yaghoobzadeh, Yadollah},
    title = "{{ParsiNLU: A Suite of Language Understanding Challenges for Persian}}",
    journal = {Transactions of the Association for Computational Linguistics},
    year = {2021},
}

@article{hendryckstest2021,
      title={{Measuring Massive Multitask Language Understanding}},
      author={Dan Hendrycks and Collin Burns and Steven Basart and Andy Zou and Mantas Mazeika and Dawn Song and Jacob Steinhardt},
      journal={Proceedings of the International Conference on Learning Representations (ICLR)},
      year={2021}
    }

@article{pnSummary,
  title={{Leveraging ParsBERT and Pretrained mT5 for Persian Abstractive Text Summarization}},
  author={Mehrdad Farahani and Mohammad Gharachorloo and M. Manthouri},
  journal={2021 26th International Computer Conference, Computer Society of Iran (CSICC)},
  year={2021},
}

@article{mukherjee2023orca,
  title={{Orca: Progressive learning from complex explanation traces of gpt-4}},
  author={Mukherjee, Subhabrata and Mitra, Arindam and Jawahar, Ganesh and Agarwal, Sahaj and Palangi, Hamid and Awadallah, Ahmed},
  journal={arXiv preprint arXiv:2306.02707},
  year={2023}
}

@article{bandarkar2023belebele,
      title={{The Belebele Benchmark: a Parallel Reading Comprehension Dataset in 122 Language Variants}}, 
      author={Lucas Bandarkar and Davis Liang and Benjamin Muller and Mikel Artetxe and Satya Narayan Shukla and Donald Husa and Naman Goyal and Abhinandan Krishnan and Luke Zettlemoyer and Madian Khabsa},
      year={2023},
      journal={arXiv preprint arXiv:2308.16884}
}

@inproceedings{xue-etal-2021-mt5,
    title = "{{mT5: A Massively Multilingual Pre-trained Text-to-Text Transformer}}",
    author = "Xue, Linting  and
      Constant, Noah  and
      Roberts, Adam  and
      Kale, Mihir  and
      Al-Rfou, Rami  and
      Siddhant, Aditya  and
      Barua, Aditya  and
      Raffel, Colin",
    booktitle = "Proceedings of the 2021 Conference of the North American Chapter of the Association for Computational Linguistics: Human Language Technologies",
    year = "2021",
}

@inproceedings{liang-etal-2023-xlm,
    title = "{{XLM-V: Overcoming the Vocabulary Bottleneck in Multilingual Masked Language Models}}",
    author = "Liang, Davis  and
      Gonen, Hila  and
      Mao, Yuning  and
      Hou, Rui  and
      Goyal, Naman  and
      Ghazvininejad, Marjan  and
      Zettlemoyer, Luke  and
      Khabsa, Madian",
    booktitle = "Proceedings of the 2023 Conference on Empirical Methods in Natural Language Processing",
    year = "2023",
}

@article{nllb2022,
  title={{No language left behind: Scaling human-centered machine translation}},
  author={Costa-juss{\`a}, Marta R and Cross, James and {\c{C}}elebi, Onur and Elbayad, Maha and Heafield, Kenneth and Heffernan, Kevin and Kalbassi, Elahe and Lam, Janice and Licht, Daniel and Maillard, Jean and others},
  journal={arXiv preprint arXiv:2207.04672},
  year={2022}
}

@inproceedings{rei-etal-2022-comet,
    title = "{{COMET-22: Unbabel-IST 2022 Submission for the Metrics Shared Task}}",
    author = "Rei, Ricardo  and
      C. de Souza, Jos{\'e} G.  and
      Alves, Duarte  and
      Zerva, Chrysoula  and
      Farinha, Ana C  and
      Glushkova, Taisiya  and
      Lavie, Alon  and
      Coheur, Luisa  and
      Martins, Andr{\'e} F. T.",
    booktitle = "Proceedings of the Seventh Conference on Machine Translation",
    year = "2022",
}

@inproceedings{schwenk-etal-2021-wikimatrix,
    title = "{{WikiMatrix: Mining 135M Parallel Sentences in 1620 Language Pairs from Wikipedia}}",
    author = "Schwenk, Holger  and
      Chaudhary, Vishrav  and
      Sun, Shuo  and
      Gong, Hongyu  and
      Guzm{\'a}n, Francisco",
    booktitle = "Proceedings of the 16th Conference of the European Chapter of the Association for Computational Linguistics",
    year = "2021",
}

@inproceedings{ghasemi2021farsick,
  title="{{FarSick: A Persian Semantic Textual Similarity And Natural Language Inference Dataset}}",
  author={Ghasemi, Zahra and Keyvanrad, Mohammad Ali},
  booktitle={International Conference on Computer Engineering and Knowledge},
  year={2021},
}

@inproceedings{muennighoff-etal-2023-mteb,
    title = "{{MTEB: Massive Text Embedding Benchmark}}",
    author = "Muennighoff, Niklas  and
      Tazi, Nouamane  and
      Magne, Loic  and
      Reimers, Nils",
    booktitle = "Proceedings of the 17th Conference of the European Chapter of the Association for Computational Linguistics",
    year = "2023",
}

@article{li2023angle,
  title="{{AnglE-optimized Text Embeddings}}",
  author={Li, Xianming and Li, Jing},
  journal={arXiv preprint arXiv:2309.12871},
  year={2023}
}

@inproceedings{devlin-etal-2019-bert,
    title = "{{BERT: Pre-training of Deep Bidirectional Transformers for Language Understanding}}",
    author = "Devlin, Jacob  and
      Chang, Ming-Wei  and
      Lee, Kenton  and
      Toutanova, Kristina",
    booktitle = "Proceedings of the 2019 Conference of the North {A}merican Chapter of the Association for Computational Linguistics: Human Language Technologies",
    year = "2019",
}

@article{ParsBERT,
  title={{ParsBERT: Transformer-based Model for Persian Language Understanding}},
  author={Farahani, Mehrdad and Gharachorloo, Mohammad and Farahani, Marzieh and Manthouri, Mohammad},
  journal={Neural Processing Letters},
  year={2021}
}

@inproceedings{feng-etal-2022-language,
    title = "{{Language-agnostic BERT Sentence Embedding}}",
    author = "Feng, Fangxiaoyu  and
      Yang, Yinfei  and
      Cer, Daniel  and
      Arivazhagan, Naveen  and
      Wang, Wei",
    booktitle = "Proceedings of the 60th Annual Meeting of the Association for Computational Linguistics",
    year = "2022",
}

@inproceedings{heffernan-etal-2022-bitext,
    title = "{{Bitext Mining Using Distilled Sentence Representations for Low-Resource Languages}}",
    author = "Heffernan, Kevin  and
      {\c{C}}elebi, Onur  and
      Schwenk, Holger",
    booktitle = "Findings of the Association for Computational Linguistics: EMNLP",
    year = "2022",
}

@article{wu2022sustainable,
  title={{Sustainable AI: Environmental implications, challenges and opportunities}},
  author={Wu, Carole-Jean and Raghavendra, Ramya and Gupta, Udit and Acun, Bilge and Ardalani, Newsha and Maeng, Kiwan and Chang, Gloria and Aga, Fiona and Huang, Jinshi and Bai, Charles and others},
  journal={Proceedings of Machine Learning and Systems},
  year={2022}
}

@article{li2023bactrian,
  title={Bactrian-X: A Multilingual Replicable Instruction-Following Model with Low-Rank Adaptation},
  author={Li, Haonan and Koto, Fajri and Wu, Minghao and Aji, Alham Fikri and Baldwin, Timothy},
  journal={arXiv preprint arXiv:2305.15011},
  year={2023}
}

@article{lialin2023scaling,
  title={Scaling down to scale up: A guide to parameter-efficient fine-tuning},
  author={Lialin, Vladislav and Deshpande, Vijeta and Rumshisky, Anna},
  journal={arXiv preprint arXiv:2303.15647},
  year={2023}
}

@misc{CarbonIntensity,
      title={{Global Electricity Review 2023}}, 
      author={Ember},
      year={2023},
      Note={\url{https://ember-climate.org/insights/research/global-electricity-review-2023/} [Accessed: 01-07-2024]}
}

@inproceedings{post-2018-call,
    title = "A Call for Clarity in Reporting {BLEU} Scores",
    author = "Post, Matt",
    booktitle = "Proceedings of the Third Conference on Machine Translation: Research Papers",
    year = "2018",
}

@inproceedings{corley2005measuring,
  title={Measuring the semantic similarity of texts},
  author={Corley, Courtney D and Mihalcea, Rada},
  booktitle={Proceedings of the ACL workshop on empirical modeling of semantic equivalence and entailment},
  pages={13--18},
  year={2005}
}

\balance

\clearpage
\appendix
\onecolumn
\pagenumbering{Roman}  %

\section{Generation samples from \pmindns}
\label{sec:appendix}
We generated the following chat discussing the enhancement of a child's math skillset.
To ensure deterministic results, we also set the \texttt{seed} value to 1442.



\vspace{1em}
\centering\includegraphics[width=0.95\textwidth]{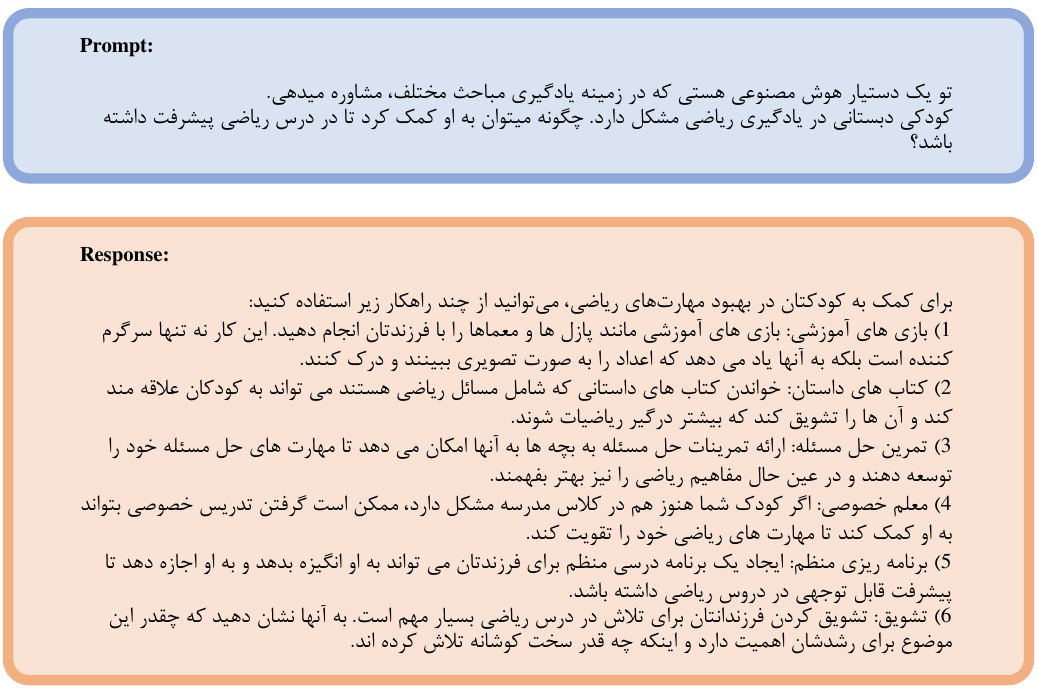}
\vspace{1em} \ \\
\centering\includegraphics[width=0.95\textwidth]{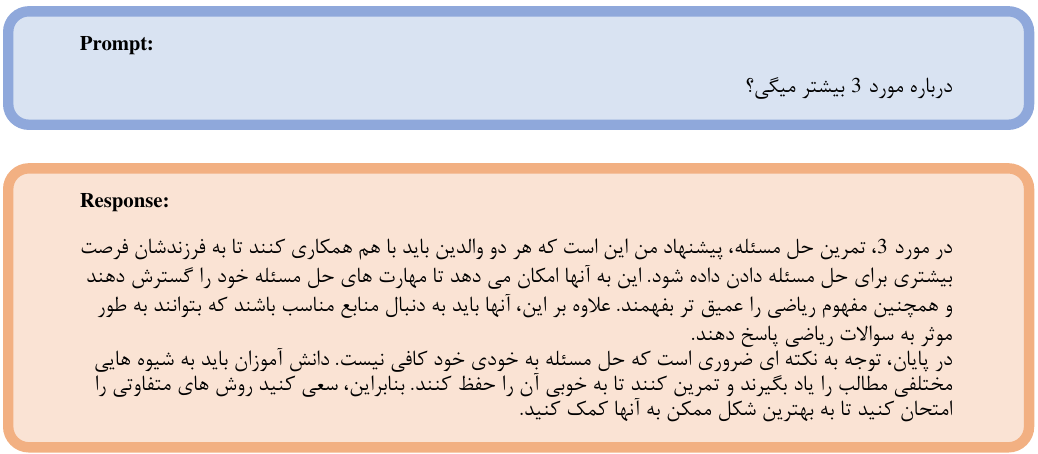}

\clearpage
\centering\includegraphics[width=0.95\textwidth]{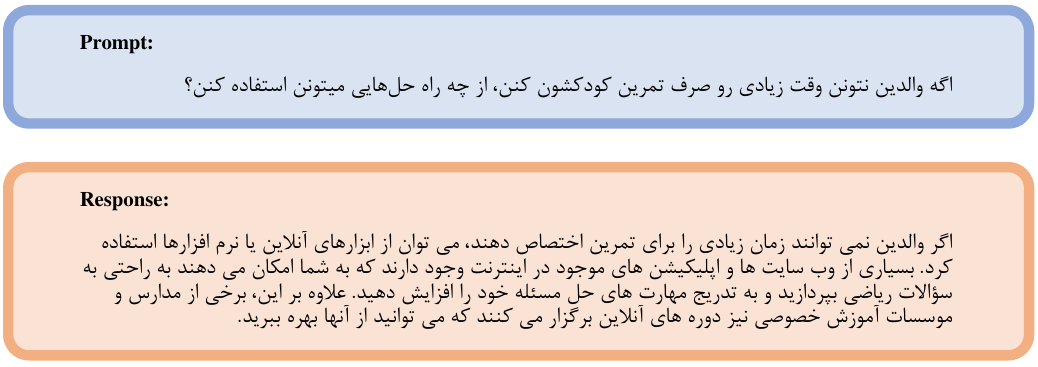}
\vspace{1em} \ \\
\centering\includegraphics[width=0.95\textwidth]{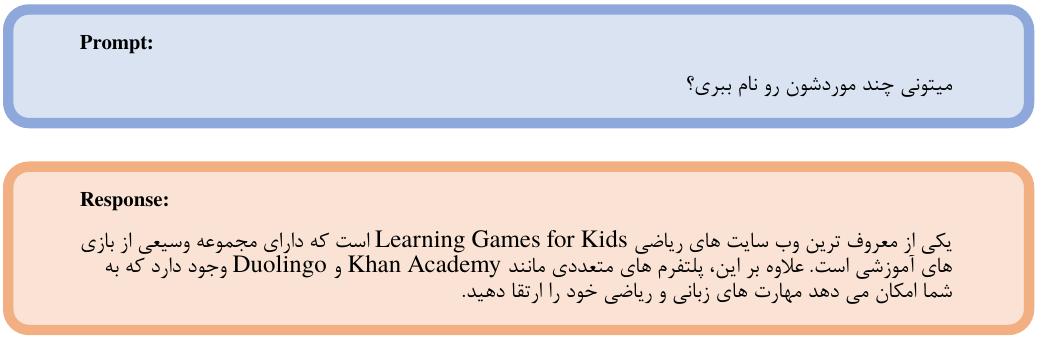}
\vspace{1em} \ \\
\centering\includegraphics[width=0.95\textwidth]{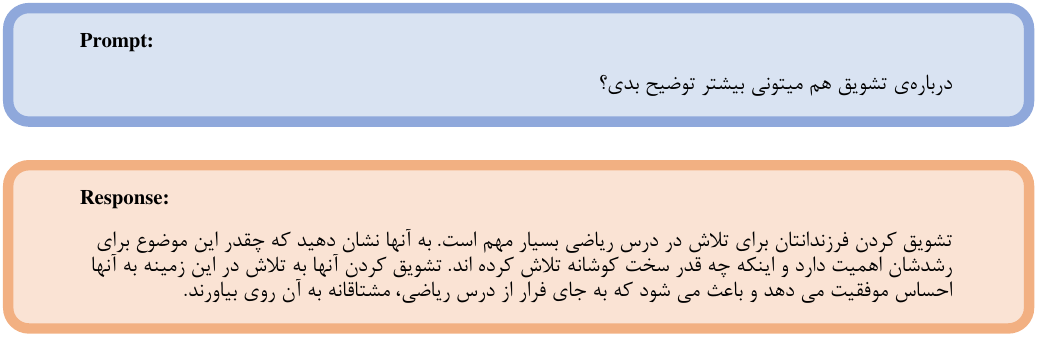}
\vspace{1em} \ \\
\centering\includegraphics[width=0.95\textwidth]{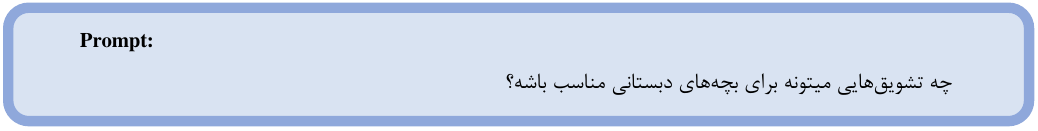}

\clearpage
\centering\includegraphics[width=0.95\textwidth]{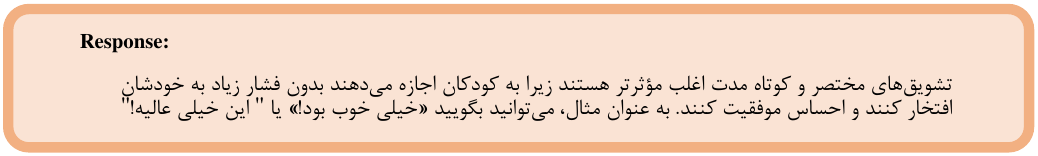}
\vspace{1em} \ \\
\centering\includegraphics[width=0.95\textwidth]{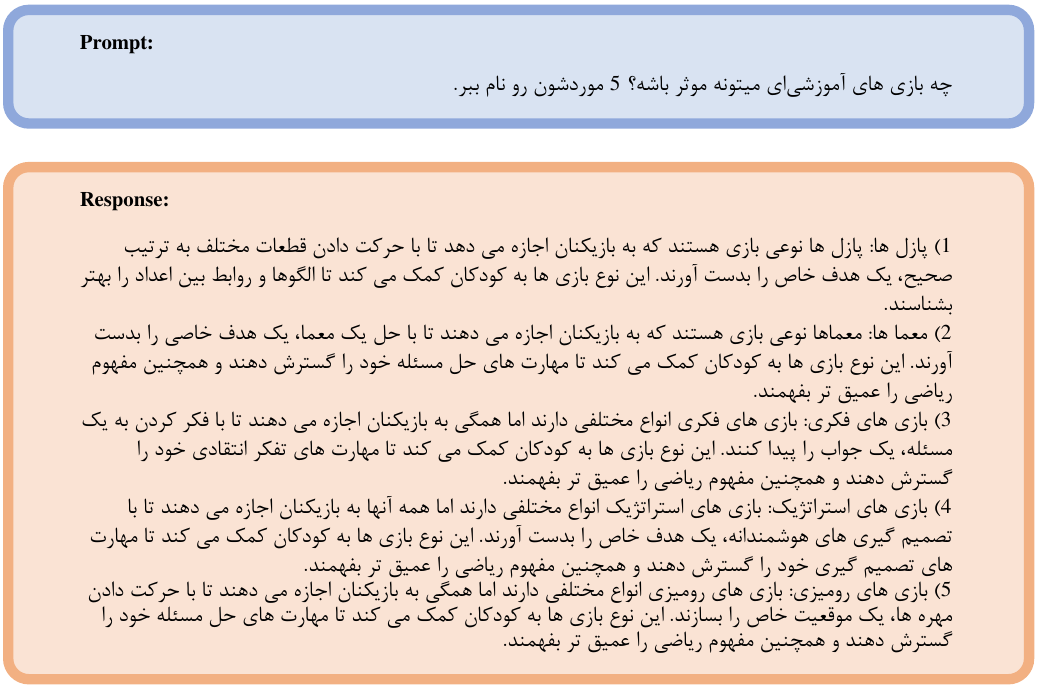}

\end{document}